\documentclass[10pt,twocolumn,letterpaper]{article} 

\usepackage{avss}
\usepackage{times}
\usepackage{epsfig}
\usepackage{graphicx}
\usepackage{amsmath}
\usepackage{amssymb}
\usepackage{caption}
\usepackage{subcaption}

\avssfinalcopy 

\def\avssPaperID{****} 

\ifavssfinal\fi

\newcommand{\seqA}{N1\textunderscore ARENA-Gp\textunderscore ENV\textunderscore RGB\textunderscore 3} 
\newcommand{\seqB}{N1\textunderscore ARENA-Gp\textunderscore TRK\textunderscore RGB\textunderscore 1} 
\newcommand{\seqC}{N1\textunderscore ARENA-Gp\textunderscore TRK\textunderscore RGB\textunderscore 2} 
\newcommand{\seqD}{W1\textunderscore P5-Gp\textunderscore TH\textunderscore 3} 
\newcommand{\seqE}{W1\textunderscore P5-Gp\textunderscore VS\textunderscore 1} 
\newcommand{\seqF}{W1\textunderscore P5-Gp\textunderscore VS\textunderscore 3} 

\begin{document}

\title{Online Pedestrian Group Walking Event Detection Using Spectral Analysis of Motion Similarity Graph  }

\author{Vahid Bastani, Damian Campo, Lucio Marcenaro and Carlo Regazzoni\\
University of Genoa, DITEN\\
Via all'Opera Pia, 11A - 16145 Genova (GE)\\
\{vahid.bastani, damian.campo\}@ginevra.dibe.unige.it, \{lucio.marcenaro, carlo.regazzoni\}@unige.it
}

\maketitle
\thispagestyle{empty}

\begin{abstract}
   A method for online identification of group of moving objects in the video is proposed in this paper. This method at each frame identifies group of tracked objects with similar local instantaneous motion pattern using spectral clustering on motion similarity graph. Then, the output of the algorithm is used to detect the event of more than two object moving together as required by PETS2015 challenge. The performance of the algorithm is evaluated on the PETS2015 dataset.
\end{abstract}

\section{Introduction}

 In the video surveillance applications, video analysis and scene understanding usually involve object detection, tracking and behavior recognition \cite{Teng2015}. A particularly important task in these applications is crowd analysis, which has been a field of great interest in computer vision and cognitive science. The crowd phenomenon has been identified as a topic of great interest in a large number of applications such as crowd management, public space design, visual surveillance and intelligent environments \cite{Zhan2008}.

Detection of events has gained lots of attention in video surveillance domain, it is the case of research related to the identification of lying pose recognition \cite{Wang2011,Volkhardt2013}, detection of people running \cite{Samuel2011,Ying2012}, crowd safety \cite{Könnecke2014501,Yin2014691}, etc. In videos that include civil interactions, modeling the social behaviors of people plays an important role in describing the individual and group behaviors on crowded scenes \cite{Yanhao2012}.

This work focuses on the understanding of events in video sequences where there are more than one person involved and groups of people walking together are conformed. Group event detection is an important application in automatic video surveillance for understanding situations that involve bunches of people doing a common activity. Identification of pedestrian groups is a key topic in crowd monitoring for the automatic recognition of anomalous behaviors that can threaten the safety of a place. For defining pedestrian groups, it is necessary to define the notion of interaction among people.

The work described in this paper studies the clustering of pedestrians into groups base on their local motion features. The output can be used for detection of events when there are pedestrian groups of more than two people in the scene. Bounding boxes represent the observation of pedestrians on each frame. The position and dimensions of bounding boxes and the speed of change of these features is tracked using Kalman filters for each object. The filtered state density of Kalman filter is used to form motion similarity graph that represent at each frame how close are the motions of pairs of pedestrians. Then, using spectral clustering techniques the group of people are identified as connected components of the graph. The proposed algorithm is tested on six videos from a dataset provided by PETS 2015 and compared with ground-truth. 

\section{Proposed Method}
The task of detecting objects moving together in the video can be viewed as determining whether they move in a same local trajectory pattern. A simple yet effective way for modeling trajectories is through flow functions \cite{Nascimento2010,Nascimento2013,Bastani2015,Kihwan2011}. In these approaches a trajectory pattern is characterized by a flow function that basically shows the speed field at each point in the environment for that pattern. Given a set of flow function corresponding to a trajectory classes, it is possible to determine the trajectory pattern class of moving objects. However, in group walking event detection task there is no flow function available beforehand. In this case the problem is to understand if the motion of two contemporary objects belong to a same flow function or not.

In this section a method is introduced for online clustering of moving pedestrians based on their motion pattern. the clustering output then is used to detect the event where more than two people walking together. To this end, first it is introduced how space-dependent flow can be estimated online for each moving object which is used for measuring the similarity of instantaneous motion pattern between objects. Then a graph will be made to represent the situation at each frame, where nodes are objects and edges are weighted by the similarity rate of motion patterns between two pair of objects. Finally, spectral graph analysis techniques are used for clustering object into groups with similar motion pattern.

\subsection{Object Detection and Tracking}
The input to the proposed algorithm at each frame is a set of bounding boxes of detected pedestrians in the scene. It is also assumed that each detection is identified using appropriate data association technique. Although this is a strong assumption, recent results on pedestrian detection \cite{Dollar2012} and multi-object tracking \cite{Bae2014} have shown its feasibility. Let $Y_t = \{\mathbf{y}^i_t, \cdots ,\mathbf{y}^{N_t}_t\}$ be the set of observations extracted from frame $t$, where $\mathbf{y}^i_t= [x , y, w, h]^T$ is the observation vector of object $i$ consisting of coordinates $(x,y)$ and dimension $(w,h)$ of its bounding box and $N_t$ is the number of objects present in frame $t$. The flow of coordinates and dimension of bonding boxes have also great  importance when we want to analyze the motion of an object. Therefore, it is reasonable to estimate  coordinates and dimension together with their flow using the sequence of observations. Kalman filter can be incorporated for this task where the state of the filter is defined as $\mathbf{x}^i_t= [x , y, w, h, \dot{x}, \dot{y}, \dot{w}, \dot{h}]^T$ consisting of coordinates and dimension of object bounding box and respective flows. 

Kalman filter at each frame provides the posterior filtered state density as a Gaussian distribution
\begin{equation}
\rho^i_t := p(\mathbf{x}^i_t|\mathbf{y}^i_{0:t}) = \mathcal{N}(\mathbf{x}^i_t; \hat{\mathbf{x}}^i_{t|t}, \hat{P}^i_{t|t})
\end{equation}
where $\hat{\mathbf{x}}^i_{t|t}$ and $\hat{P}^i_{t|t}$ are the filtered state and its covariance matrix up to time $t$. Note that the state of Kalman filter in this case consists of position and flow of all four components. Each instance of the state vector is one sample from trajectory pattern flow function. For this reason we use the posterior state of Kalman filter to compute the similarity of the motion of objects at each frame.

\subsection{Motion Pattern Similarity Measure}
it is possible to use the euclidean distance between the estimated state vector of each object to measure their similarity. However, since the output of Kalman filter has the form of probability distribution, it is more effective to use distance metrics specific for probability distribution. Kullback-Leibler (KL) divergence is a measure of difference between probability distributions denoted by $D_{KL}(P||Q)$. It is available in closed form for Gaussian distributions \cite{Hershey2007}. However, KL is not a symmetric measure since $D_{KL}(P||Q) \neq D_{KL}(Q||P)$. Thus, here the symmetrized version of KL divergence is used to measure differences of two state distribution of moving objects:
\begin{equation}
d^{i,j}_t := D_{KL}(\rho^i_t||\rho^j_t)/2 + D_{KL}(\rho^j_t||\rho^i_t)/2.
\end{equation}

$d^{i,j}_t$ is a positive value that increases as the difference between  $\rho^i_t$ and $\rho^i_t$ becomes larger. A normalized similarity between two state vector distribution then is calculated as
\begin{equation}
s^{i,j}_t = e^{-d^{i,j}_t/k^{i,j}_t}.
\label{eq:sim}
\end{equation}
$k^{i,j}_t$ is a scaling factor whose motivation is to compensate the effect of distance of objects from camera. Note that as objects move away from camera their speed and respective distance get lower in the image plane. Thus a global measure of similarity should be equalized as far as possible in order to give the same score for the same situation whether it is close or far from camera. The scaling factor is defined as
\begin{equation}
k^{i,j}_t = a(\sqrt{w^i_t h^i_t}+\sqrt{w^j_t h^j_t})/2+b.
\label{eq:scale}
\end{equation}
Which is a linear function of the mean of the square root of the area of two objects. Here the area of the objects is used to understand the relative distance of them from camera. $a$ and $b$ both are positive so that the scaling factor increases as the objects become close to camera. In the similarity measure (\ref{eq:sim}), the scaling factor compensates the larger relative distance of objects when they are close to camera.

\begin{figure}[t]
\begin{center}
   \includegraphics[width=0.6\linewidth]{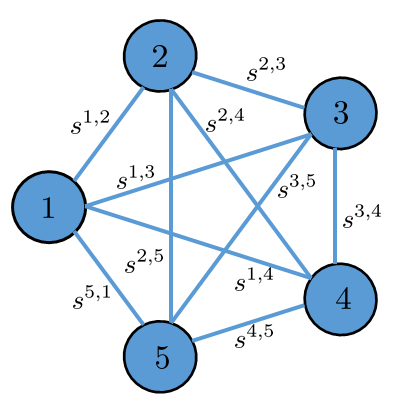}
\end{center}
   \caption{Example of similarity graph for 5 moving objects.}
\label{fig:graph}
\end{figure}

 $s^{i,j}_t$ is one when  $\rho^i_t$ and $\rho^i_t$ are completely similar and monotonically decreases toward zero as they become more different. The pairwise similarity scores then are used to form similarity graph such as the one shown in the Fig.\ref{fig:graph}, which is an undirected graph whose nodes represent moving objects and edges are weighted according their motion similarity. The graph later will be used to understand relations between objects and detection of the event when more than two people walking together.

\subsection{Group Walking Event Detection}
The spectral clustering algorithm \cite{Malik2000} is a simple yet effective algorithm for clustering data sets that can be represented using similarity graphs. It often outperforms many conventional algorithms. In the proposed algorithm we use spectral clustering on the generated motion similarity graph in order to find groups of objects moving in a same way. The adjacency matrix for a graph such as Fig.\ref{fig:graph} is the matrix $W_t = [s^{i,j}_t]_{i,j = 1,\cdots,N}$ \cite{Luxburg2007}. The main quantity for spectral clustering is graph Laplacian matrix which is defined as
\begin{equation}
L_t = D_t-W_t.
\end{equation}
where $D_t$ is the graph degree matrix  defined as a diagonal matrix whose $i$th diagonal element is the $i$th node degree $m^i_t = \sum_{j=1}^n s^{i,j}_t$. The Laplacian matrix can be used to find connected components of the graph. In the application of this paper the connected components of the graph represent groups of people walking together.

Let $\lambda_1 \leq \lambda_2 \leq \cdots \lambda_n$ be the eigenvalues of Laplacian matrix $L_t$ and $\mathbf{v}_1, \mathbf{v}_2, \cdots, \mathbf{v}_n$ be corresponding  eigenvectors, e.g. $L_t \mathbf{v}_i = \lambda_i  \mathbf{v}_i$ for $i=1,\cdots,n$. If the number of connected components $m \leq n$ of the graph is known, the connected components and their corresponding nodes can be found using spectral clustering algorithm \cite{Malik2000} as follows:
\begin{itemize}
\itemsep-.1em 
  \item Let $U$ be the matrix whose columns are the first $m$ eigenvectors $\mathbf{v}_1, \cdots, \mathbf{v}_m$.
  \item For $i=1,\cdots,n$, let $\mathbf{u}_i$ be the vector corresponding to $i$th row of $U$.
  \item Cluster vectors $\{\mathbf{u}_i\}_{i=1,\cdots,n}$ into $m$ clusters using $k$-means algorithm into clusters $C_1,\cdots,C_m$.
  \item Return cluster indicator variables $z^i_t = \{j | \mathbf{u}_i \in C_j\}$ for $i=1,\cdots,n$.
\end{itemize}
The event of a group of people walking is then triggered when the number of members in at least one connected component (cluster) is greater than or equal to three.

\begin{figure}[t]
\begin{center}
   \includegraphics[width=0.9\linewidth]{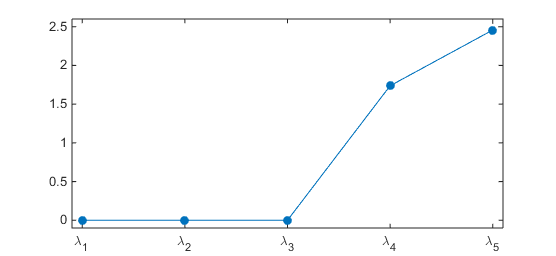}
\end{center}
   \caption{Example of Laplacian matrix eigenvalues for a graph with 5 nodes and 3 connected components. From motion similarity graph of frame 47 of sequence \seqA.}
\label{fig:eig}
\end{figure}

The number of connected components (the number of groups of people in scene) however is not known in this problem. A particular way to estimate number of connected components is eigngap heuristic \cite{Luxburg2007}. Here $m$ is chosen such that all eigenvalues $\lambda_1, \cdots, \lambda_m$ are small and $\lambda_{m+1}$ is relatively large. This is shown in the example of Fig.\ref{fig:eig} where three first three eigenvalues are close to zero and there is a big gap from third to forth eigenvalues, which shows that there are three connected components in the corresponding graph. Thus, to find $m$ using eigngap heuristic the following procedure is done:
 \begin{itemize}
\itemsep-.1em 
  \item Calculate $\gamma_i = \lambda_{i+1}-\lambda_{i}$ for $i=1,\cdots,n-1$
  \item Find $m= \min i$ such that $\gamma_i \geq \frac{0.8}{n} \sum_{j=1}^{n-1} \gamma_j$ 
\end{itemize}
The above procedure searches for the first gap in the sequence of eigenvalues . Since the first eigenvalue is always zero \cite{Luxburg2007}, the first gap shows the number of connected components in the graph. 

\section{Experimental results}
The PETS2015 dataset\footnote{http://pets2015.net./} is used to evaluate the proposed algorithm. The data sets consist of three different camera views of two situations P5 and ARENA, in which a group of walking people appears in the scene. In ARENA sequences a group of three people appear in an environment in which there are some other individual pedestrians. This group eventually splits into three separate pedestrians. In P5 sequences, a group of six people are shown which eventually splits into one individual, a group of two and a group of three people while walking.

The detection of pedestrians done manually using Viper-GT tool\footnote{http://viper-toolkit.sourceforge.net/}. The bounding boxes of objects then are feed sequentially to the algorithm. The id of objects are also passed to algorithm in order to bypass data association problem. For evaluation purpose a ground-truth of  group indexes of objects in each frame of video is made by human observer. The ground-truth group (cluster) indicator variables denoted by $g^i_t$ is compared with the algorithm output $z^i_t$. Since the problem is formulated here as a clustering problem, it is possible to use clustering performance measures such as Adjacent Mutual Information (AMI) \cite{Vinh2010} to evaluate the algorithm. Given the ground-truth of the cluster assignments this metric compares the result of clustering by returning values in range [0 1]. AMI Values close to zero means two assignments are highly independent and AMI values close to one indicates match between two indexes. The AMI score is calculated at each frame $t$ by comparing $\{g^i_t\}_{i=1}^n$ with $\{z^i_t\}_{i=1}^n$. The mean AMI value for every frame of each sequence then is reported here.

\begin{figure*}
\begin{center}
	\begin{subfigure}[b]{0.3\textwidth}
                \includegraphics[width=\textwidth]{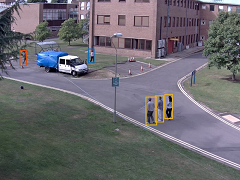}
                \caption{\seqA}
                \label{fig:gull}
        \end{subfigure}
	\begin{subfigure}[b]{0.3\textwidth}
                \includegraphics[width=\textwidth]{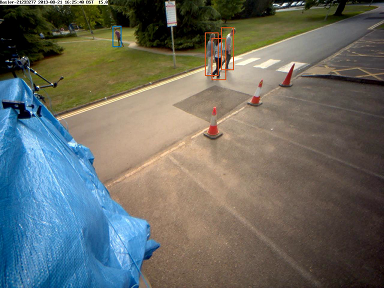}
                \caption{\seqB}
                \label{fig:gull}
        \end{subfigure}
	\begin{subfigure}[b]{0.3\textwidth}
                \includegraphics[width=\textwidth]{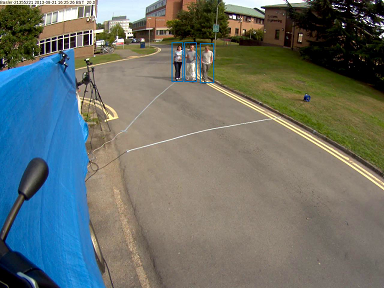}
                \caption{\seqC}
                \label{fig:gull}
        \end{subfigure}
	\begin{subfigure}[b]{0.3\textwidth}
                \includegraphics[width=\textwidth]{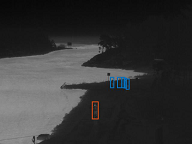}
                \caption{\seqD}
                \label{fig:gull}
        \end{subfigure}
	\begin{subfigure}[b]{0.3\textwidth}
                \includegraphics[width=\textwidth]{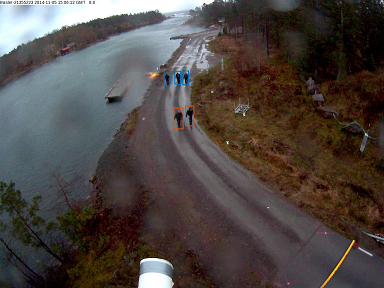}
                \caption{\seqE}
                \label{fig:gull}
        \end{subfigure}
	\begin{subfigure}[b]{0.3\textwidth}
                \includegraphics[width=\textwidth]{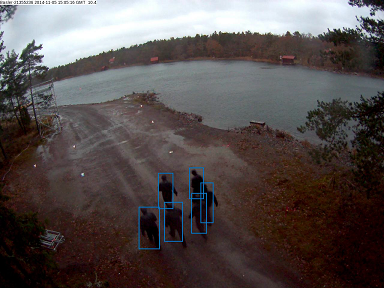}
                \caption{\seqF}
                \label{fig:gull}
        \end{subfigure}
\end{center}
   \caption{Snapshots of the output of the algorithm from P5 and ARENA sequences. Same color boxes represent same group.}
\label{fig:snap}
\end{figure*}

\begin{figure}
\begin{center}
	\begin{subfigure}[b]{0.4\textwidth}
                \includegraphics[width=\textwidth]{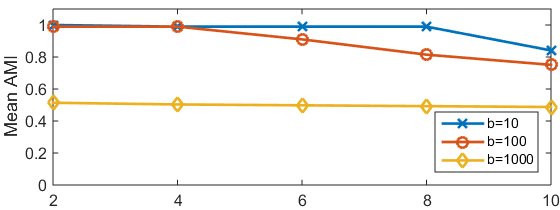}
                \caption{\seqA}
                \label{fig:gull}
        \end{subfigure}
	\begin{subfigure}[b]{0.4\textwidth}
                \includegraphics[width=\textwidth]{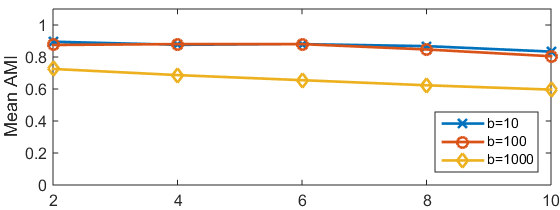}
                \caption{\seqB}
                \label{fig:gull}
        \end{subfigure}
	\begin{subfigure}[b]{0.4\textwidth}
                \includegraphics[width=\textwidth]{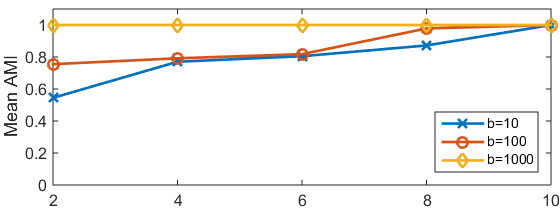}
                \caption{\seqC}
                \label{fig:gull}
        \end{subfigure}
	\begin{subfigure}[b]{0.4\textwidth}
                \includegraphics[width=\textwidth]{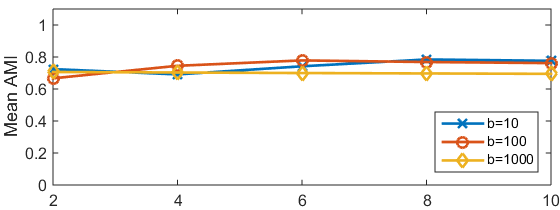}
                \caption{\seqD}
                \label{fig:gull}
        \end{subfigure}
	\begin{subfigure}[b]{0.4\textwidth}
                \includegraphics[width=\textwidth]{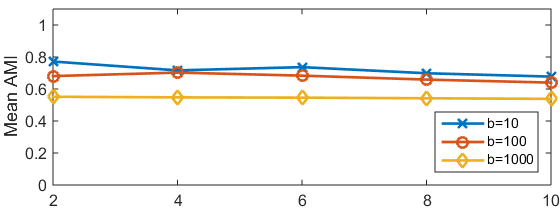}
                \caption{\seqE}
                \label{fig:gull}
        \end{subfigure}
	\begin{subfigure}[b]{0.4\textwidth}
                \includegraphics[width=\textwidth]{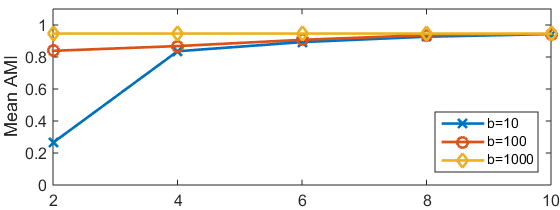}
                \caption{\seqF}
                \label{fig:gull}
        \end{subfigure}
\end{center}
   \caption{Mean AMI value versus different choices of parameters $a$ (horizontal axis) and $b$ for six sequences from P5 and ARENA dataset of PETS2015.}
\label{fig:res}
\end{figure}

The measurement noise covariance matrix in Kalman filter is set to $10\times \mathbf{I}_4$, and the process noise covariance matrix is set to
\begin{equation*}
\begin{bmatrix}
       10\times \mathbf{I}_4 & \mathbf{0} \\[0.3em]
       \mathbf{0} & 2\times \mathbf{I}_4\\[0.3em]
     \end{bmatrix}.
\end{equation*}
The algorithm is evaluated for different choices of parameters $a$ and $b$ of (\ref{eq:scale}) to see how its behavior changes with different parameter setting. Parameter $a$ is chosen from set $\{2, 4, 6, 8, 10\}$ and  parameter $b$ is chosen from set $\{10, 100, 100\}$. Fig \ref{fig:snap} shows snapshots of the output of the algorithm where objects that are in same group have same color bounding boxes. The quantitative results are depicted in Fig. \ref{fig:res} where for each sequence the value of mean AMI is plotted versus the value of parameter $a$ for different value  parameter $b$. As can be observed from Fig. \ref{fig:res} the performance depends on the situation, camera view and parameters setting. However, the performances in any case is reasonably good since the value of mean AMI is close to one. The best average AMI score over all sequences is 0.8566, which is achieved for parameter set $a=8$ and $b=10$.

\section{Conclusions}
In this paper a method for online clustering of walking people is proposed for detecting of the event when more than two people are walking together. The method is based on measuring the similarity of the motion patterns of pairs of moving objects in the scene to form a motion similarity graph. The graph then is used to cluster objects based on spectral clustering algorithm. Experimental results show that the proposed method is able to identify separate walking groups efficiently. This method however is instantaneous and does not take into account the history of objects. An improvement can be achieved by applying probabilistic filtering like Hidden Markov Model (HMM) on the output group indexes to eliminate sporadic joining and splitting of groups.

{\small
\bibliographystyle{ieee}
\bibliography{egbib}
}


\end{document}